\documentclass[10pt,twocolumn,letterpaper]{article}

\usepackage[pagenumbers]{cvpr} 

\usepackage{graphicx}
\usepackage{amsmath}
\usepackage{amssymb}
\usepackage{booktabs}
\usepackage{multirow}
\usepackage{dsfont}
\usepackage{pifont}% http://ctan.org/pkg/pifont
\newcommand{\cmark}{\ding{51}}%
\newcommand{\xmark}{\ding{55}}%

\usepackage[pagebackref,breaklinks,colorlinks]{hyperref}

\usepackage[capitalize]{cleveref}
\crefname{section}{Sec.}{Secs.}
\Crefname{section}{Section}{Sections}
\Crefname{table}{Table}{Tables}
\crefname{table}{Tab.}{Tabs.}

\begin{document}

\title{Giga-SSL: Self-Supervised Learning for Gigapixel Images}
\author{Tristan Lazard \\
CBIO, Mines Paris, PSL University \\
Paris, France \\
{\tt\small tristan.lazard@mines-paristech.fr}
\and
Marvin Lerousseau\\
CBIO, Mines Paris, PSL University \\
Paris, France\\
{\tt\small marvin.lerousseau@mines-paristech.fr}
\and
Etienne Decencière\\
CMM, Mines Paris, PSL University \\
Fontainebleau, France \\
{\tt\small etienne.decenciere@mines-paristech.fr}
\and 
Thomas Walter\\
CBIO, Mines Paris, PSL University \\
Paris, France\\
{\tt\small thomas.walter@mines-paristech.fr}
}
\maketitle

%%%%%%%%% ABSTRACT
\begin{abstract}
    Whole slide images (WSI) are microscopy images of stained tissue slides routinely prepared for diagnosis and treatment selection in medical practice. WSI are very large (gigapixel size) and complex (made of up to millions of cells). The current state-of-the-art (SoTA) approach to classify WSI subdivides them into tiles, encodes them by pre-trained networks and applies Multiple Instance Learning (MIL) to train for specific downstream tasks. However, annotated datasets are often small, typically a few hundred to a few thousand WSI, which may cause overfitting and underperforming models. Conversely, the number of unannotated WSI is ever increasing, with datasets of tens of thousands (soon to be millions) of images available. While it has been previously proposed to use these unannotated data to identify suitable tile representations by self-supervised learning (SSL), downstream classification tasks still require full supervision because parts of the MIL architecture is not trained during tile level SSL pre-training. Here, we propose a strategy of slide level SSL to leverage the large number of WSI without annotations to infer powerful slide representations. Applying our method to The Cancer-Genome Atlas, one of the most widely used data resources in cancer research (16 TB image data), we are able to downsize the dataset to 23 MB without any loss in predictive power: we show that a linear classifier trained on top of these embeddings maintains or improves previous SoTA performances on various benchmark WSI classification tasks. Finally, we observe that training a classifier on these representations with tiny datasets (\eg 50 slides) improved performances over SoTA by an average of +6.3 AUC points over all downstream tasks. 

\end{abstract}

\section{Introduction}
\label{sec:intro}

Whole slide images (WSI) are microscopy images of stained tissue sections. They are enormous (billions of pixels) and complex, often containing millions of individual cells, their environments, and the overall tissue structure. They are routinely used in cancer treatment centers for diagnosis, patient stratification, and treatment selection. Computational pathology is the field concerned with the automatic analysis of WSI. The most clinically impactful task in computational pathology is to make predictions directly from the WSI, such as predicting cancer subtype, survival of the patient, or response to treatment.
The major challenges in building predictive models operating on WSI are:

\begin{itemize}
\item Prohibitive memory requirements (typically 15GB uncompressed per WSI);
\item Signal/noise: The high amount of biological material, not necessarily related to the output variable, is making models: (i) fail to identify the region of interests; (ii) prone to overfitting.
\item Technical complexity: WSI are technically demanding to deal with given their large size, which presents a considerable barrier for multi-modal analyses of genomic and pathology data.
\end{itemize}

Today, the leading methods for WSI classification rely on Multiple Instance Learning (MIL): WSI are tessellated into small images, called tiles, which are encoded by an embedder. Tile embedders are usually pre-trained, either on natural images or - more recently and with great effect - by self-supervised learning (SSL). WSI are then seen as bags of tiles, and the slide representation is obtained by combining the tile embeddings, which are then used as input for the slide classification network. The agglomeration strategy comes in different flavors and usually relies on tile selection or weighted averaging of tile embeddings \cite{courtiol_deep_2019,ilse_attention-based_2018,lu_data-efficient_2021,rymarczyk_kernel_2020,li_dual-stream_2020}. The slide classification network is usually trained from scratch on the specific classification task. 

While these methods successfully predict a large variety of output variables, such as grade, cancer subtype, gene signatures, mutations or response to treatment \cite{campanella_clinical-grade_2019,coudray_classification_2018,kather_pan-cancer_2020,lazard_deep_2021,naylor_neural_2022,echle_deep_2021,qu_genetic_2021}, the performances remain highly dependent on the size of the training dataset \cite{campanella_clinical-grade_2019}. Indeed, MIL performance reaches saturation when using thousands of slides with associated ground truth for training \cite{campanella_clinical-grade_2019}. This might be realistic for the most frequent cancer types and routinely acquired output variables, but in most real-world projects only a few tens or hundreds of WSI with corresponding ground truth are available. However, with the digitalization of many pathology facilities, there is an increasing access to WSI without ground truth which are digitalized in clinical routine. Following the SSL paradigm that has been successfully applied at the tile level \cite{dehaene_self-supervision_2020,lazard_deep_2021,ciga_self_2021,saillard_self_2021}, there is a challenging opportunity to make use of these unannotated data at the slide level to derive meaningful slide representations. These would be particularly useful for small cohorts and non-standard output variables, such as prognosis for rare cancer types or prediction of treatment response in clinical trials.

However, learning representations at the WSI level is difficult since WSI cannot be manipulated as one image object due to their size, impeding the straightforward use of self-supervised learning frameworks developed on natural images.
The community needs to innovate to translate SSL at the WSI level regarding the design of pertinent augmentations.
For instance, the crop augmentation plays a central role for learning good representations with SSL on natural images \cite{chen_simple_2020,misra_self-supervised_2019}.
However, randomly cropping one memory-fittable image from a WSI can lead to a complete loss of the cells and tissues that determine its ground-truth, due to the inherent heterogeneity of tissues.
Further developments should also be done on the architecture of a SSL framework for WSI representations, as was done in the only paper tackling SSL at the WSI level \cite{chen_scaling_2022}.

Here, we propose Giga-SSL, a strategy to perform SSL for gigapixel images. Designed for pathology data, our method is capable of leveraging large datasets, such as The Cancer Genome Atlas (TCGA) \cite{weinstein_cancer_2013}, to learn representations at the WSI level without using any ground truth data -- but only whole slide images. 
Our main contributions are:

\begin{itemize}
    \item Giga-SSL, an efficient self-supervised learning framework for gigapixel images.
    \item Extensive experiments show that a linear classifier that uses these embeddings outperforms the current state-of-the-art performance on several clinically impactful classification tasks. The gains are especially significant for small datasets.
    \item We publicly release the WSI embeddings of the whole TCGA dataset, compressing it by a factor of almost 1 million from 16Tb to 23Mb, and thus making this large image datasets amenable for future research.
\end{itemize}

We expect that this method will have an important impact in the field of computational pathology in two ways: (1) Our method specifically boosts performance for small datasets, which are very common in practice. We therefore address a major bottleneck in computational pathology. (2) 
We can make image data accessible to a larger community of researchers in cancer bioinformatics, in order to investigate the complex relationships between genetic, transcriptomic and phenotypic data. 
To facilitate reproducibility and the broad use of Giga-SSL, the complete source code of this work as well as the full TCGA-FFPE encodings are available at https://github.com/trislaz/gigassl.

\section{Background}

\subsection{Multiple instance learning for gigapixel images}
In the MIL paradigm, objects (called bags) comprise other objects (called instances). 
For gigapixel images, the bag is a gigapixel image, and its instances are subimages (also called tiles or patches) extracted throughout the gigapixel image. 
While traditional MIL assumes independent and identically distributed (i.i.d.) instances within each bag \cite{ilse_attention-based_2018}, this assumption is relaxed for gigapixel images because instances are extracted from the same image, and are therefore not independent. 
Given a gigapixel image $X$ made of $n_x$ instances $(x_1, \dots, x_{n_x})$, MIL is implemented as a combination of three modules: (i) an instance embedder $e_{\theta_1}(\cdot)$, (ii) a pooling operator $p_{\theta_2}(\cdot)$ and (iii) a classifier $c_{\theta_3}(\cdot)$ such that a decision $\hat{y}$ is obtained with

\begin{equation*}
\hat{y} = c_{\theta_3}\Big( p_{\theta_2}\big(\{e_{\theta_1}(x_1), \dots, e_{\theta_1}(x_n)\}\big) \Big).
\end{equation*}

Most  MIL architectures differ in the design of the pooling operator $p_{\theta_2}$. 
There are two families of operators: (i) those that consider instances as i.i.d. and (ii) those that exploit the relationship between instances of a bag. 
Architectures that consider instances as i.i.d. are either parameterless (\eg using the operators average, maximum, a concatenation of both \cite{lerousseau_multimodal_2020}, or a noisy-OR function \cite{srinivas_generalization_2013}), or trainable, such as an attention-based neural network \cite{ilse_attention-based_2018}. 
While these architectures obtain good performances, instances of gigapixel images are dependent and contain information that can be leveraged to produce accurate predictions. 
Modern MIL architecture for gigapixel images have been designed to exploit the spatial relationship of instances. 
For instance, transformer-based MIL approaches \cite{shao_transmil_2021} extend the attention mechanism of Ilse \etal \cite{ilse_attention-based_2018} by incorporating the positions of instances for decision prediction. 
Of particular interest in this work, the SparseConvMIL \cite{lerousseau_sparseconvmil_2021} architecture leverages spatial information by building a sparse map from both the instance embeddings and their sampled locations. 
This map is further processed by a sparse-input convolutional neural network that outputs a latent vector to be further classified by a generic classifier.

\begin{figure*}[t!]
  \centering
  \includegraphics[width=\linewidth]{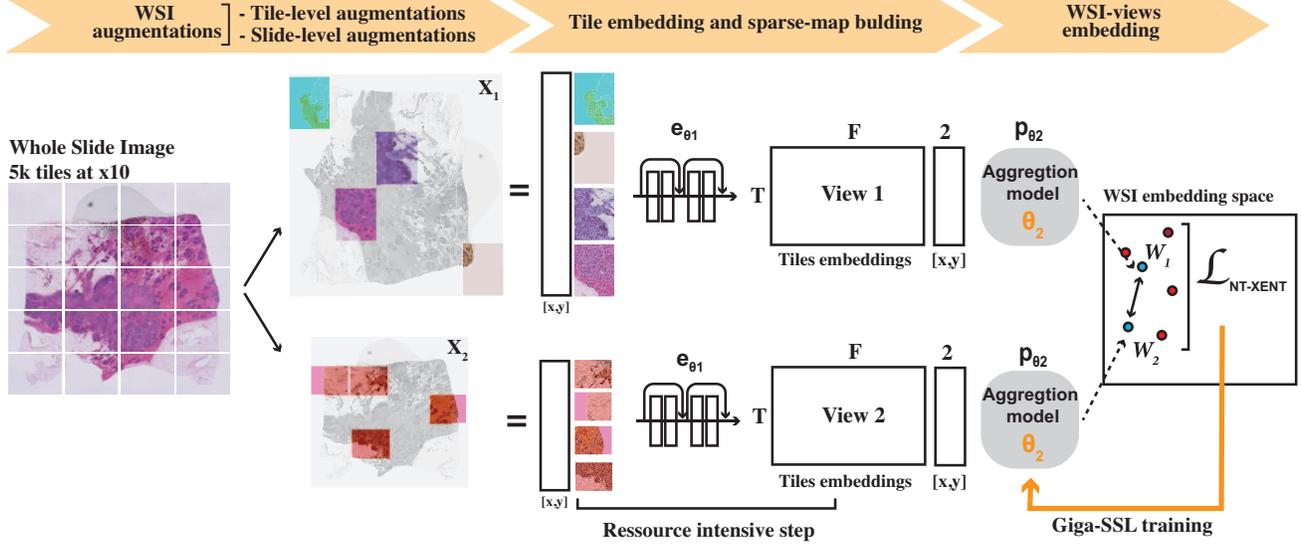}
  \caption{Overview of the Giga-SSL method. First, random augmentations of a WSI are used to create two different views $X_1$ and $X_2$ of the same WSI. 
  Next, $T$ tiles randomly extracted from each view are embedded using a tile-embedder network $e_{\theta_1}$, resulting in $T$ embeddings in $\mathbb{R}^F$. 
  These embeddings and their associated tile coordinates are fed into a sparse-input CNN model $p_{\theta_2}$, producing two WSI representations $W_1$ and $W_2$. 
  A contrastive loss is applied on a minibatch of several whole slide images in order to update both $e_{\theta_1}$ and $p_{\theta_2}$.
  }
  \label{fig:architecture}
\end{figure*}

\subsection{Self-supervised learning for gigapixel images}
Self-supervised learning have been investigated in computational pathology at the tile level, \ie for patches extracted from whole slide images \cite{dehaene_self-supervision_2020,lazard_deep_2021,ciga_self_2021,saillard_self_2021}.
The findings suggest that SSL indeed improved the performance on WSI classification tasks by using the SSL pre-trained tile level model as a frozen tile encoder.
Because patches extracted from WSI are of size similar to datasets of natural images, the majority of the work successfully used off-the-shelf frameworks developed on natural images such as SimCLR \cite{chen_simple_2020} or MoCo \cite{he_momentum_2020}.

To the best of our knowledge, only one prior work has proposed a self-supervised learning framework for learning representations directly at the WSI level \cite{chen_scaling_2022}.
To do so, the authors design a new architecture made of 3 hierarchically stacked visual transformers \cite{dosovitskiy_image_2020} which is trained on unlabelled WSI with the DINO framework \cite{caron_emerging_2021}, notably by enforcing consistency between two perturbed views of the same object.
As stated by the authors \cite{chen_scaling_2022}, their approach cannot be trained end-to-end due to memory issues and needs to be trained in stages, starting from the visual transformer at higher magnification.
on top of time-consuming SSL pre-training, a drawback is the need to retrain all transformers at lower magnifications when modifying one visual transformer.
A major bottleneck of this approach is the necessity to retrain the last transformer from scratch% since the authors did not manage to train it with more than 10,000 WSI
, implying that (i) the whole system does not benefit fully of SSL pretraining, and that (ii) linear embeddings cannot be extracted for new slides and used as input vectors for downstream tasks \cite{chen_scaling_2022}.
Conversely, we designed an efficient method for learning WSI representations that obtained state-of-the-art performance with a linear classifier without the need to fine-tune any part of our system.

\section{Methods}

\subsection{Algorithmic design}
\paragraph{Notations and algorithmic background}  \label{notations}
Giga-SSL training comprises 6 sequential steps to extract WSI representations which we details here and which is illustrated in Figure \cref{fig:architecture}.
Lets us consider a WSI $X$.
Giga-SSL uses an extension of the SparseConvMIL architecture for WSI classification \cite{lerousseau_sparseconvmil_2021} by considering a ResNet network $f_\theta$ (\eg ResNet18) \cite{he_deep_2015}, which is cut at the beginning of the fourth residual block into two sequential parts:
\begin{enumerate}
    \item the first part, acting as the tile embedder $e_{\theta_1}$, is made of all layers of $f_\theta$ up to the first layer of the fourth block, %(\ie all pre-block layers including the first $7\times7$ convolution, the first, second and third residual block),
    \item the second part, acting as the pooling function $p_{\theta_2}$, is made of all layers after and including the fourth block of $f_\theta$,
\end{enumerate}
such that for any image $i$, the ResNet embedding is:
$$
f_\theta(i) = p_{\theta_2}\big( e_{\theta_1}(i) \big) \in \mathbb{R}^{512}.
$$

\paragraph{Step 1: Augmentation of the WSI at the tile-level} 
\label{methods_step1}
Two augmentation functions $t_1$ and $t_2$ are sampled from an image augmentation domain $A$ made of color augmentations (color jitter, grayscale) and geometric augmentations (flips, rotations, scaling, blurring).
First, $T$ tiles are subsampled from $X$ for each augmentation function $t_1$ and $t_2$, yielding two sets of patches $\{X_1\}$ and $\{X_2\}$.
The coordinates of the top-left pixel of the tiles are stored for further processing.
Finally $t_1$ is applied to all patches of $\{X_1\}$, yielding a set of augmented patches denoted as $t_1\big(\{X_1\}\big)$, and similarly a set $t_2\big(\{X_2\}\big)$ for the second set patches $\{X_2\}$.

\paragraph{Step 2: Embedding of tiles} 
\label{methods_step2}
Each tile of both $t_1\big(\{X_1\}\big)$ and $t_2\big(\{X_2\}\big)$ are concurrently and independently forwarded through the tile embedder network $e_{\theta_1}$.
Each image is thus converted into a feature map which is averaged across all pixels, yielding a tile embedding of size $F$ (256 for ResNet18) for each tile of $t_1\big(\{X_1\}\big)$ and $t_2\big(\{X_2\}\big)$

\paragraph{Step 3: Building of the sparse maps}
\label{methods_step3}
Following the framework of SparseConvMIL \cite{lerousseau_sparseconvmil_2021}, a sparse map $S_1$ is built by assigning each produced embedding of $t_1\big(\{X_1\}\big)$ at the location where each of its original tiles was sampled in Step 1 \cref{methods_step1} but downsampled by a factor $d=224$.
Similarly, a sparse map $S_2$ is built from the embeddings $t_2\big(\{X_2\}\big)$.

\paragraph{Step 4: Augmentations of the WSI at the slide-level}
While WSI are difficult to manipulate due to their huge size, a sparse map can be augmented with geometric transformations, enabling our framework to perform slide-level transformations in real-time.
$S_1$ and $S_2$ are randomly flipped, rotated, and scaled with a factor uniformly sampled in $[0.5, 2]$ independently for the $x$ and $y$ axis.

\paragraph{Step 5: Embedding of the sparse maps into two augmented WSI representations}
\label{methods_step5}
To compute representations, we apply $p_{\theta_2}$ on both augmented sparse maps $S_1$ and $S_2$.
It should be noted that $p_{\theta_2}$ is not a conventional CNN model but has been converted into a submanifold
convolutional network \cite{graham_submanifold_2017} with the same architecture such that it can process sparse data.
At this stage, the two augmented views of the input WSI $X$ (augmented at the tile-level and at the slide-level) are vector representations of the WSI. 

\paragraph{Step 6: Loss optimization}
As is done in SimCLR, augmented views are finally fed to a projector, giving two augmented projections with which the loss will be computed.
We train the weights of the pooling function $p_{\theta_2}$ by optimizing the contrastive loss NT-XENT loss \cite{chen_simple_2020}. 
Given a minibatch $B$ of augmented WSI $(X_1^{i}, X_2^{i})_{i \in B}$, we set the loss function for a positive pair of WSI as
\begin{equation}
   \ell_{i}=-\log \frac{\exp \left(\operatorname{sim}\left(X_1^{i}, X_2^{i}\right) / \tau\right)}{\sum_{x \in B} \mathbf{1}_{\{x \neq X_1^{i}\}} \exp \left(\operatorname{sim}\left(X_1^{i}, x\right) / \tau\right)}
\end{equation}
where $\tau$ is the temperature parameter and $\mathbf{1}_{\{.\}}$ the indicator function. The final loss is computed as the average of these terms across all views. %  $$controlling the sharpness of the produced ground-truth.

\subsection{Design choices}

\paragraph{Selection of the underlying CNN architecture and loss function}
Giga-SSL does not theoretically rely on a ResNet architecture.
There are many choices of good architectures that could be used for the comprising tile encoder and pooling function, including two parts of different architectures.
However, the pooling function must be implemented such that it can handle sparse data since it processes the augmented sparse maps (see Step 5 \cref{methods_step5}).

\paragraph{Freezing the tile encoder} \label{frozen}
A key computational bottleneck of this strategy is the online computation of tile embeddings for a batch of $B$ WSI, each composed of $T$ tiles.% of size 256x256.
GPU memory limitations put constraints on $B$ and $N_t$, which effectively limits the number of total tiles per batch that can be used.
Besides, it has been shown in SSL for natural images that a large batch size is required to yield representations with good downstream classification performances \cite{chen_exploring_2020,chen_simple_2020,chen_intriguing_2021}.
A strategy for overcoming these issues is to freeze the tile encoder $e_{\theta_1}$ and pre-compute the embeddings of randomly sampled and augmented tiles for each WSI, \ie essentially bypassing steps 1 and 2 of \cref{methods_step2}.
For encoding a WSI, this is implemented by:
(i) sampling 50 tile-level augmentation functions (both color and geometric augmentations) $(t_k)_{k \leqslant 50}$,
(ii) for each $k$, randomly subsampling 256 tiles from the WSI and augment them with $t_k$,
and (iii) concurrently and independently forwarding each augmented tile into $e_{\theta_1}$ and storing them.
This process leads to $N$*50*256 tile embeddings where $N$ is the total number of WSI of the Giga-SSL training dataset.

Giga-SSL is then trained, starting from step 3 \cref{methods_step3} by performing the following to sample a view of a WSI:
(i) sample one of the 50 tile-level augmentations,
(ii) sample a subset $T$ of the 256 embeddings obtained from this augmentation,
(iii) build the sparse map,
and (iv) carry on from step 4 of \cref{methods_step1}.

\section{Experimental validation}

\subsection{Step 1: self-supervised pre-training}

Self-supervised pre-training of Giga-SSL is done using The Cancer Genome Atlas (TCGA) \cite{weinstein_cancer_2013}, a public dataset that comprises 11754 whole slide images containing  tissue from virtually all types of solid cancers. This dataset is the result of an international data-collecting effort and therefore features a high variety of participant centers (190).
Such slides are crucial for patient care since they are the basis of diagnosis and treatment selection. 
On average, images have a width of 93000 pixels and a height 67500 pixels, for an average of 6.5 billion  pixels per image. 
Fully compressed, TCGA weighs more than 16 Terabytes, \ie 3 orders of magnitude more than ImageNet \cite{deng_imagenet_2009}. 
We tesselated non-overlapping square patches of size 256 pixels from all diagnostic slides of the TCGA at 10x magnification.

\textbf{$e_{\theta_1}$ pre-training} We choose to pre-train $e_{\theta_1}$ using MoCo \cite{he_momentum_2020}.
We trained a full ResNet18 on a subset of 6 million of these tiles extracted from a random set of 3000 slides from the TCGA for 200 epochs. $e_{\theta_1}$ is then extracted from this network as described in \cref{notations}. More details about this pre-training are available in the supplementaries.

\textbf{Giga-SSL pretraining:} we trained Giga-SSL on the full TCGA dataset, with frozen augmented embeddings extracted with the previously described pre-trained tile embedder (see \cref{frozen}), with Adam \cite{kingma_adam:_2014} for 1000 epochs.

\subsection{Step 2: learning from linear embeddings}

\paragraph{Training design} 
For Giga-SSL, similarly to the works on natural images \cite{chen_simple_2020,he_momentum_2020,caron_emerging_2021}, we measured the quality of the learned representations by performing linear probing either with all the labels available for a given task or by artificially reducing the number of labels to simulate a semi-supervised setting. 
To do so, one representation was extracted for each WSI after SSL pretraining. 
These representations were then used as input data to train a logistic regression for each considered downstream task.

\paragraph{Datasets} This protocol was applied to six diagnostic WSI classification tasks highly pertinent for clinical practice: 

\begin{itemize}
    \item 3 tasks performed by Chen \etal \cite{chen_scaling_2022} aiming at automating the routine diagnosis of Non-Small Scell Lung Cancer (NSCLC), Breast Cancer (BRCA), and Kidney Cancer (RCC);
    \item 3 tasks aiming at inferring molecular properties from tissue slides towards faster, cheaper and more accessible molecular testing for cancer therapy selection.
\end{itemize}
For each of these 6 tasks, \cref{experiments_labels_per_class} reports the number of training WSI of the corresponding dataset, and their class distribution.
All the datasets for these tasks are subsets of the TCGA \cite{weinstein_cancer_2013}.
Results were computed on 10 bootstrapped splits of the data for each experiment, as was done in Chen \etal \cite{chen_scaling_2022}, and we also used their train/test splits to ensure fairness of performance comparisons.

\begin{table}[!htp]\centering
\begin{tabular}{lrrr}\toprule
Task&\# samples &\# labels per class \\\midrule
BRCA subtyping &1041 &831 - 210 \\
Kidney subtyping &924 &510 - 294 - 120 \\
NSCLC subtyping &1033 &528 - 505 \\
BRCA Molecular &595 &129 - 466 \\
BRCA mHRD &912 &447 - 465 \\
BRCA tHRD &634 &318 - 316 \\
\bottomrule
\end{tabular}
\caption{Total number of samples and number of samples per class for all of the 6 benchmarked tasks in this paper.}
\label{experiments_labels_per_class}
\end{table}

\begin{table*}[!t]\centering
\scalebox{0.9}{
\begin{tabular}{lccccccc}\toprule
&Method &Giga-SSL (proposed)  &AverageMIL &DeepMIL\cite{ilse_attention-based_2018} &HIPT\cite{chen_scaling_2022} &DeepSMILE\cite{schirris_deepsmile_2021} \\\cmidrule{2-7}\cmidrule{5-7}
&Linear &\cmark &\cmark &\xmark &\xmark &\xmark \\\cmidrule{2-7}
Task &\% data & & & & & \\\midrule
\multirow{2}{*}{$\text{NSCLC}_{\text{subtyping}}$} &100 &\textbf{0.952} $\pm$ 0.020 &0.913 $\pm$ 0.023 &0.948 $\pm$ 0.017 &\textbf{0.952} $\pm$ 0.021&- \\
&25 &\textbf{0.939} $\pm$ 0.017 &0.885 $\pm$ 0.036&0.922 $\pm$  0.034 &0.923 $\pm$ 0.020 &- \\ \hline
\multirow{2}{*}{$\text{BRCA}_{\text{subtyping}}$} &100 &\textbf{0.905} $\pm$ 0.032 &0.859 $\pm$ 0.038&0.874 $\pm$ 0.050&0.874 $\pm$ 0.060&- \\
&25 &\textbf{0.890} $\pm$ 0.058&0.822 $\pm$ 0.072  &0.860 $\pm$ 0.042 &0.821 $\pm$ 0.069&- \\\hline
\multirow{2}{*}{$\text{RCC}_{\text{subtyping}}$} &100 &0.982 $\pm$ 0.007&0.973 $\pm$ 0.011 &\textbf{0.986} $\pm$ 0.008&0.980 $\pm$ 0.013&- \\
&25 &\textbf{0.975} $\pm$ 0.012&0.959 $\pm$ 0.015 &0.970 $\pm$ 0.016&0.974 $\pm$ 0.012 &- \\\hline
\multirow{2}{*}{$\text{BRCA}_{\text{molecular}}$} &100 &\textbf{0.938} $\pm$ 0.035 &0.920 $\pm$ 0.037&0.924 $\pm$  0.042&- &- \\
&25 &\textbf{0.853} $\pm$ 0.075 &0.799 $\pm$ 0.068 &0.810 $\pm$ 0.093&- &- \\\hline
\multirow{2}{*}{BRCA mHRD } &100 &\textbf{0.756} $\pm$ 0.028 &0.706 $\pm$ 0.030 &0.736 $\pm$ 0.047&- &0.727 $\pm$ 0.010 \\
&25 &\textbf{0.743} $\pm$ 0.039 &0.643 $\pm$ 0.050 &0.660 $\pm$ 0.046 &- &- \\\hline
\multirow{2}{*}{BRCA tHRD} &100 &\textbf{0.855} $\pm$ 0.023 &0.799 $\pm$ 0.034 &0.836 $\pm$ 0.052 &- &0.838 $\pm$  0.012\\
&25 &\textbf{0.781} $\pm$ 0.050 &0.698  $\pm$ 0.078 &0.721  $\pm$ 0.075 &- &- \\
\bottomrule
\end{tabular}}
\caption{Benchmark study reporting the 10-fold cross-validated AUC performances of a logistic regression trained with Giga-SSL WSI representations or AverageMIL WSI representations, and retrained from scratch for other benchmarked approaches. For each task, we evaluate the methods with two data budgets with either 100\% or 25\% of the available training data.\label{results}}

\end{table*}

\paragraph{Default settings}
The number $T$ of tiles sampled per slide to 5.
For a slide $X$, we bootstrap $R=50$ views without tile augmentation (\ie differing only in the sampled tiles), compute their embedding $\{W_r\}_{1, \dots, 50}$ and consider the WSI representation as the elementwise average of the $\{W_r\}_{1, \dots, 50}$.
Average embeddings are normalized using a standard scale, while the Giga-SSL embeddings are normalized using the L2 unit.
More details about training parameters are given in the supplementaries.

\subsection{Results}

\paragraph{Classification results on benchmarked tasks}
\Cref{results} synthesizes the results on all tasks for 5 models \ie average, an attention-based MIL \cite{ilse_attention-based_2018} on top of a ResNet18 pretrained with MoCo , DeepSMILE \cite{schirris_deepsmile_2021} and HIPT\cite{chen_scaling_2022}. Results from HIPT and DeepSMILE are taken from their respective articles, and constitute the SoTA on the task on which they are cited.

%%%
Our proposed approach, Giga-SSL, outperforms the state-of-the-art on two out of three tasks benchmarked in (\cite{chen_scaling_2022}) when using 100\% of the available training labels NSCLC and BRCA subtyping. For BRCA subtyping, the AUC is increased by 3 points. Our proposed approach also achieves superior performances for all the other remaining tasks (mHRD, tHRD and BRCA molecular profiling). However, the power of the proposed approach seems to be in the low data regime. This is evident by the results obtained by using only 25\% of the available labels. In this semi-supervised regime, the proposed approach obtained the best results on all tasks. 
While this finding may be expected when comparing Giga-SSL to methods without pretraining,  Giga-SSL obtained superior results compared to the other SSL-based approach HIPT. For example, there is a gain of 6.9 AUC  points for BRCA subtyping.

Compared to attention-based MIL and HIPT, the proposed approach (Giga-SSL) provides an overall gain in performance while working in a linear regime. This is in contrast to HIPT and attention-based methods, which require fine-tuning and learning from scratch, respectively. Consequently, the downstream training pipeline for Giga-SSL is extremely efficient in comparison to the other two approaches. For instance, training for BRCA subtyping with 100\% of the training data on 10 bootstrapped splits took 1.25 CPU-seconds for the proposed approach versus 150 GPU-minutes for attention-based MIL. This is a difference of 7200 times in favor of Giga-SSL – while also obtaining superior performances.

\paragraph{Tiny datasets} 
In practice, pathological datasets can be tiny for the prediction of treatment response.
For instance, phase II clinical trials typically involve 50 patients. 
Training a model to identify responding and non-responding patients is therefore challenging due to the low number of available labels. 

We measured the performance of Giga-SSL in such a context by artificially reducing the size of all 6 datasets to 250, 100 and 50 samples.
We compare Giga-SSL to the DeepAttnMIL model, which performances are on par with all other benchmarked algorithms (see \cref{results}).

\begin{figure*}[h]
    \begin{center}
    \includegraphics[scale=0.6]{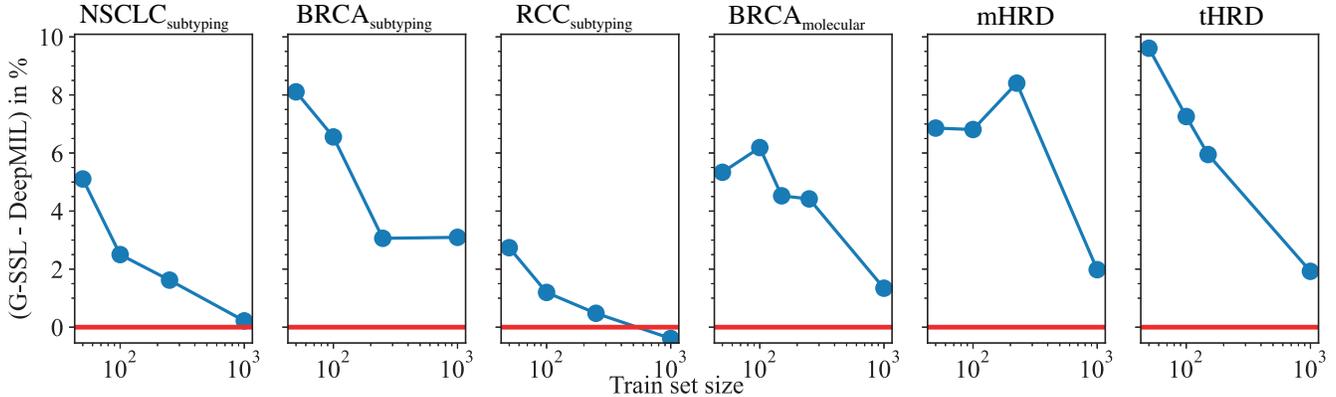}
    \end{center}
    \caption{Difference between the average AUC performances of Giga-SSL and DeepMIL (in \%) as a function of the training set size. The red line represents equal performance. Above the red line, the advantage is given to Giga-SSL \label{tiny}}
\end{figure*}

\Cref{tiny} shows that the performance gap between the proposed approach and the standard WSI classification method strengthens as the number of samples decreases.
The average improvement over all tasks brought by Giga-SSL features is of 5.1 AUC points when using 100 WSI and up to 6.3 AUC points when using only 50 WSI.

\section{Ablation study and sensitivity analyses}
In this section, we aim to understand the impact of some of Giga-SSL design choices over the predictive power of the learned representations.
All subsequent experiments were conducted with the same conditions (including hyperparameters, epochs, and training dataset) as in the previous experiments, unless otherwise stated. 

\paragraph{Sharing tile augmentations within views improves performance}
\Cref{ablation} reports the performance of Giga-SSL when removing one component at a time, \ie (i) with a tile embedder pre-trained on ImageNet rather than pre-trained with MoCo on histopathological data ($\text{Giga-SSL}_{\text{im}}$), (ii) without slide-level augmentation during the WSI-level SSL pretraining; (iii) without shared augmentations across all tiles of a view, \ie each tile is transformed by a randomly and independently sampled augmentation.

\begin{table}[!htp]\centering
\scriptsize
\begin{tabular}{lrrrrrrr}\toprule
&\multicolumn{3}{c}{100\% data} &\multicolumn{3}{c}{50 WSI} \\\cmidrule{2-7}
&NSCLC &CRC &BRCA &NSCLC &CRC &BRCA \\\midrule
Giga-SSL &0.952 &0.982 &0.905 &0.894 &0.960 &0.793 \\
w/o slide-aug &0.935 &0.973 &0.894 &0.86 &0.951 &0.80 \\
NS &0.933 &0.971 &0.875 &0.847 &0.939 &0.774 \\
$\text{Giga-SSL}_{\text{im}}$ &0.922 &0.978 &0.888 &0.813 &0.952 &0.751 \\
$\text{Giga-SSL}_{\text{im}}$ NS  &0.897 &0.975 &0.853 &0.777 &0.935 &0.707 \\
\bottomrule
\end{tabular}
\caption{\label{ablation} 10-fold cross-validated AUC performances of ablated Giga-SSL models. w/o slide-aug is a Giga-SSL model trained without slide-level augmentations. NS (Not Shared) is a Giga-SSL model trained without sharing the tile-level augmentation among views. $\text{Giga-SSL}_{\text{im}}$ stands for a Giga-SSL model trained with tiles embeddings transfered from an ImageNet pretraining.}
\end{table}

Using a tile-level SSL algorithm to pretrain the tile encoder $e_{\theta_1}$ brings improvement to the WSI-level representations: the Giga-SSL trained with MoCo features outperforms its ImageNet ($\text{Giga-SSL}_{\text{im}}$) counter part on all tasks.
On the contrary, the slide-level augmentation does not seem to be extremely important for the SSL task, as removing it has a small to no impact on performances.

However, applying independent transformations to each tile (\textit{not shared}) degrades substantially the performances with an average decrease of 1.9 AUC points using 100\% of the data down to 2.8 AUC points when using only 50 WSI, over the classification tasks. 
When ablating the shared transformations from a Giga-SSL model trained with tile features pretrained with ImageNet, the drop of performances compared to a $\text{Giga-SSL}_{\text{im}}$ is even more important: 2.1 AUC points with 100\% of the data, 3.2 AUC points with 50 WSI.

Using shared augmentation thus allows the learning of useful features in abundant and scarce data regimes. 
We hypothesize key features linked to the slide preparation and shared by all the tiles on the slide are still available for shortcut learning if the tile-level augmentations are not shared. It seems that these shortcut features may be more present in ImageNet than in MoCo.
Highlighting such features and finding even more stringent ways to hide them when learning Giga-SSL would improve even more its performances.

\begin{figure*}[h]
\begin{center}
\includegraphics[scale=0.7]{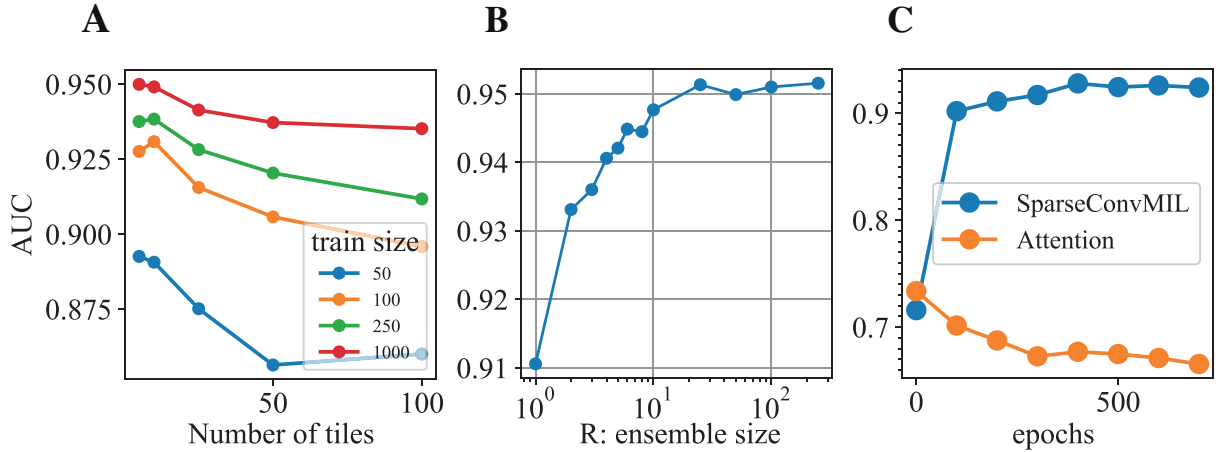}
\end{center}
\caption{\label{res2}Experiments on key parameters of Giga-SSL. Each point is a 10-fold cross-validated AUC performance of a logistic regression fed with Giga-SSL features. The classification task is NSCLC subtyping for the three experiments. \textbf{A}. Effect of the number of sampled tiles $T$ per WSI during training. \textbf{B. }Effect of the number $R$ of bootstrapped non-augmented views of WSI to feed Giga-SSL at inference time (see  supplementaries).\textbf{ C.} Evolution of the performances of a Giga-SSL with a SparseConvMIL (blue line, normal situation) or an attention-MIL network (orange line) as an aggregator.}
\end{figure*}

\paragraph{The fewer tiles, the better}
\Cref{res2}.A presents the performances of 4 Giga-SSL models trained with different numbers of sampled tiles per view. The fewer tiles we sample, the better the resulting WSI representations. This behaviour strengthens when the downstream problem has a smaller training set and is comparable among all the downstream classification tasks.
Interestingly, we can observe the opposite effect when using a DeepMIL model to classify a WSI: the fewer tiles used at training time, the worse the performances \cite{lerousseau_multimodal_2020}.
A very small number $T$ of sampled tiles per view when training Giga-SSL can be seen as an aggressive augmentation.
It has been reported (\cite{chen_simple_2020}) that SSL benefits from stronger augmentations more than classification tasks, and Tian \etal (\cite{tian_what_2020}) have shown that there is an optimal strength of augmentation for each downstream task. This optimum results from a trade-off between keeping enough information to solve the downstream task and minimizing irrelevant features.

As sampling 5 tiles per WSI is enough to learn useful information to solve all the proposed downstream tasks, we can deduce that the signal relative to these problems is distributed among most of the tiles of the WSI.
It would be interesting to test the performances of Giga-SSL on a classification task for which we know that the signal is highly concentrated on a few instances.

\paragraph{Ensembling representations brings improvement}
We show in supplementaries that a Giga-SSL model with a SparseConvMIL aggregation module must use the same number of tiles per WSI at inference and training. We therefore decided to bootstrap $R$ views of a WSI at inference time before averaging the Giga-SSL embeddings of these $R$ views.
Figure \cref{res2}.B investigates the effect of $R$ on the downstream performances of the Giga-SSL representations. It shows that without this ensembling strategy, Giga-SSL loses up to 4 AUC points on NSCLC subtyping. The gain in performance saturates around $R=50$.

\paragraph{Generalization} \label{generalization} Giga-SSL has been trained on the full TCGA dataset, and downstream classification dataset also comes from the TCGA. In order to investigate the extent to which Giga-SSL could transfer to other datasets, we extracted from the TCGA all slides coming from the 41 centers that contributed to the NSCLC dataset, leading to an independent set of  6840 WSI. We trained Giga-SSL for 1000 epochs on this training set and reports the results in table \cref{tab: generalization}.
\begin{table}[!htp]\centering
\scalebox{0.9}{
\begin{tabular}{lrrr}\toprule
data regime &100\% data &50 WSI \\\midrule
Full dataset &0.952 $\pm$ 0.020 &0.894 $\pm$ 0.045\\
Independent training set &0.948 $\pm$ 0.017 &0.885 $\pm$ 0.045 \\
\bottomrule
\end{tabular}}
\caption{Linear classification performances on NSCLC subtyping of embeddings trained on either the full TCGA or a subset of the TCGA independent from the downstream task dataset.}\label{tab: generalization}
\end{table}
Interestingly, Giga-SSL performs almost as good when trained on a set of WSI totally independent from the downstream task set. This suggests that Giga-SSL would generalize well on a different dataset.

\paragraph{Attention-deep-MIL unlearns when trained with SSL}
Instead of using a sparse-CNN as a tiles features aggregator, one could choose any other MIL model. We trained a Giga-SSL model with a DeepMIL aggregation module and evaluated its downstream linear performances on the NSCLC dataset.
\Cref{res2}.C shows that the performances of such a model decrease while the SSL training is in progress.
Although the DeepMIL shows very good classification performances \cref{results} when trained from scratch, this architecture seems not suitable for Giga-SSL pretraining. 
We suspect that the DeepMIL architecture has too easily access to shortcuts features to learn the WSI identity. Understanding what causes its collapse may highlight key pitfall for Giga-SSL training and therefore allow to improve it.

For all of the latter points, we report in the supplementaries a similar behaviour on the other downstream classification tasks.

\section{Conclusion} 
\textbf{Limitations} While Giga-SSL has been shown to generalize well outside of its training data distribution, the tile-embedder is not pre-trained on a dataset that is entirely independent from the downstream tasks datasets. It would be interesting to conduct the same experiment as \cref{generalization} but excluding the WSI from the tile-embedder pre-training dataset too. 
In addition, a drawback of working with frozen embeddings of WSI is that it removes any possibility of building explainable models.

\textbf{Finally,} we have explored self-supervised learning for whole slide images with a versatile design based on specific data augmentation tailored for the multiple instance learning framework. 
Our proposed approach achieved or beat state-of-the-art performance over a wide range of clinically impactful tasks in both high and low data regimes. 
In particular, for small datasets (\eg 50 slides), our approach achieved a performance improvement of 6.3 AUC points on average compared to competing methods. 
Ablation studies and sensitivity analyses highlighted the key components of our approach – including tile encoder pretraining and how to apply augmentations to tiles – to better understand the pitfalls of self-supervised whole slide image representation learning.

The public release of the learned representations for all diagnostic slides of The Cancer Genome Atlas in a manageable size has the potential to decipher new knowledge about cancer and to develop new tools for diagnosis assistance and treatment response prediction towards improved patient survival. 

{\small
\bibliographystyle{ieee_fullname}
\bibliography{references}
}
\end{document}